\documentclass[sigconf]{acmart}
\AtBeginDocument{%
  }

\usepackage{algorithm}
\usepackage{algpseudocode}

\usepackage{multirow}

\usepackage{makecell}

\usepackage{float}
\usepackage{adjustbox}
\usepackage{caption}
\usepackage{cuted} 
\usepackage{listings}
\usepackage{xcolor}

\lstdefinestyle{prompt}{
    basicstyle=\ttfamily\scriptsize,
    breaklines=true,
    breakatwhitespace=true,
    backgroundcolor=\color{gray!5},
    frame=single,
    framerule=0.3pt,
    columns=fullflexible,
    keepspaces=true,
    showstringspaces=false,
    aboveskip=4pt,
    belowskip=4pt,
    tabsize=2,
    lineskip=-1pt
}

\acmConference{ }
\acmDOI{}
\acmISBN{}
\renewcommand\footnotetextcopyrightpermission[1]{}
\begin{document}

\title{From Rows to Reasoning: A Retrieval-Augmented Multimodal Framework for Spreadsheet Understanding}

\author{Anmol Gulati}
\affiliation{%
  \institution{Commercial Technology and Innovation Office, PricewaterhouseCoopers}
  \country{U.S.}
}

\author{Sahil Sen}
\affiliation{%
  \institution{Commercial Technology and Innovation Office, PricewaterhouseCoopers}
  \country{U.S.}
}

\author{Waqar Sarguroh}
\affiliation{%
  \institution{Commercial Technology and Innovation Office, PricewaterhouseCoopers}
  \country{U.S.}
}

\author{Kevin Paul}
\affiliation{%
  \institution{Commercial Technology and Innovation Office, PricewaterhouseCoopers}
  \country{U.S.}
}

\renewcommand{\shortauthors}{Gulati et al.}

\begin{abstract}
Large Language Models (LLMs) struggle to reason over large-scale enterprise spreadsheets containing thousands of numeric rows, multiple linked sheets, and embedded visual content such as charts and receipts. Prior state-of-the-art spreadsheet reasoning approaches typically rely on single-sheet compression or full-context encoding, which limits scalability and fails to reflect how real users interact with complex, multimodal workbooks. We introduce FRTR-Bench, the first large-scale benchmark for multimodal spreadsheet reasoning, comprising 30 enterprise-grade Excel workbooks spanning nearly four million cells and more than 50 embedded images. To address these challenges, we present From Rows to Reasoning (FRTR), an advanced, multimodal retrieval-augmented generation framework that decomposes Excel workbooks into granular row, column, and block embeddings, employs hybrid lexical-dense retrieval with Reciprocal Rank Fusion (RRF), and integrates multimodal embeddings to reason over both numerical and visual information. We tested FRTR on six LLMs, achieving 74\% answer accuracy on FRTR-Bench with Claude Sonnet 4.5, a substantial improvement over prior state-of-the-art approaches that reached only 24\%. On the SpreadsheetLLM benchmark, FRTR achieved 87\% accuracy with GPT-5 while reducing token usage by roughly 50\% compared to direct serialization methods.
\end{abstract}


\maketitle

\section{Introduction}

Spreadsheets remain the foundational medium for analytical work at an enterprise level, and modern workbooks routinely contain thousands of cells, span multiple sheets, and embed diverse media such as charts, receipts, and scanned tables \cite{kandel2012enterprise, hermans2015enron}. While Large Language Models (LLMs) promise intuitive natural-language interaction with such documents, directly serializing entire sheets into a prompt is inefficient and brittle. Spreadsheet semantics hinge on spatial layout, structural dependencies, and cross-sheet relationships rather than text alone. Moreover, recent work shows that long-context LLMs suffer positional degradation, as accuracy declines when relevant information appears far from the prompt boundaries \cite{liu2023lostmiddlelanguagemodels}. This motivates retrieval-focused approaches that surface small, relevant subsets rather than streaming full spreadsheets into the context window.

\begin{figure*}[t]
    \centering
    \includegraphics[width=.8\textwidth]{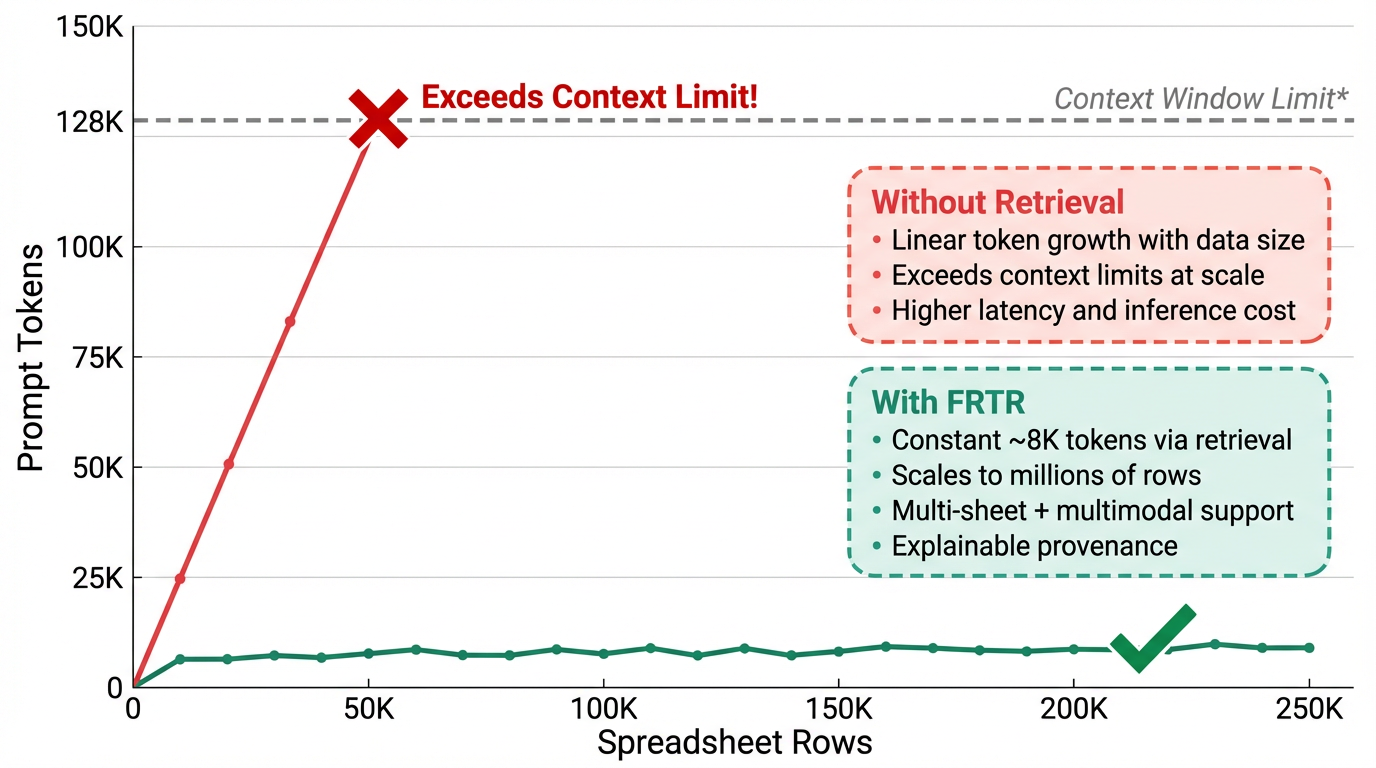}
    \Description{Scalability comparison showing prompt token usage vs spreadsheet rows. Without retrieval, tokens grow linearly and exceed context limits. With FRTR, tokens remain constant at approximately 8K regardless of data size.}
    \caption{Token scaling comparison: naive full-context serialization exceeds context limits as data grows, while FRTR maintains constant token usage ($\sim$8K) via selective retrieval, enabling scalable enterprise reasoning.}
    \label{fig:scalability}
\end{figure*}

A prominent direction is SpreadsheetLLM \cite{dong2025spreadsheetllmencodingspreadsheetslarge}, which introduces encoding schemes SheetEncoder and SheetCompressor that compress cell content, positional addresses, and formatting to fit within an LLM's context window. While effective for single-sheet reasoning, such compression assumes all relevant data can fit within a single prompt and fails to scale to multi-sheet or multimodal settings. In practice, enterprise workbooks exceed these boundaries, incorporating thousands of rows, inter-sheet references, and embedded images. Retrieval-augmented generation approaches provide a complementary path: combining database-style retrieval with hybrid lexical–dense retrieval to fetch only the minimal, most relevant evidence for reasoning \cite{cormack2009rrf, karpukhin2020dpr, khattab2020colbert}. Empirical studies demonstrate that retrieval-augmented generation (RAG) can match or surpass long-context models at significantly lower computational cost and latency \cite{lewis2021retrievalaugmentedgenerationknowledgeintensivenlp, gao2024retrievalaugmentedgenerationlargelanguage,borgeaud2022retro, karpukhin2020dpr}. Simple rank fusion methods, such as Reciprocal Rank Fusion (RRF), further enhance stability and performance across retrieval modalities \cite{azureHybridSearchRRF}.

In this work, we make three novel contributions:

\begin{itemize}
    \item \textbf{Retrieval-first framework:} We present an advanced, multimodal retrieval-augmented generation approach to spreadsheet understanding that integrates hybrid lexical–dense retrieval search to overcome full-context limitations and improve scalability.
    
    \item \textbf{Multi-granular and multimodal indexing:} We introduce a fine-grained indexing scheme for Excel workbooks at the row, column, and block levels, and extend it to multimodal settings using joint text–vision embeddings. Hybrid rank-fusion retrieval ensures robust and reproducible performance across both textual and image-bearing spreadsheets.
    
    \item \textbf{FRTR-Bench benchmark:} We release \textit{FRTR-Bench}, the first large-scale, multimodal spreadsheet reasoning benchmark, spanning 30 enterprise-grade workbooks with nearly four million cells and over fifty embedded images, to support reproducible evaluation of retrieval-augmented spreadsheet understanding.
\end{itemize}

By reframing spreadsheet analysis as a three-part pipeline composed of retrieval, verification, and composition, FRTR incorporates information-retrieval principles, emphasizing minimal access and explicit evidence grounding for LLM-spreadsheet interaction. The result is a scalable, auditable, and cost-efficient framework for reasoning in finance, analytics, and auditing.

\section{Related Work}

\subsection{LLMs for Tables and Spreadsheets}

Early neural models such as TAPAS \cite{herzig2020tapas} learned to answer questions directly over tables by encoding cell values jointly with queries, while TUTA \cite{yang2022tuta} extended this paradigm by modeling hierarchical and spatial cues. OmniTab \cite{deng2022omnitab} further improved few-shot table QA by combining natural and synthetic supervision. These approaches focus on \emph{in-context} reasoning within a single table and do not scale to multi-sheet or multimodal workbooks.

More recent work such as \textit{SpreadsheetLLM} \cite{dong2025spreadsheetllmencodingspreadsheetslarge} compresses spreadsheet structure and content to fit within LLM context limits, achieving notable gains on single-sheet reasoning. Complementary methods, including Table-GPT \cite{cheng2023tablegpt} and TableLLM \cite{chen2023tablellm}, improve general table manipulation and code generation capabilities. However, these approaches are not capable of handling cross-sheet and multimodal reasoning. In contrast, FRTR emphasizes hybrid lexical-dense retrieval and multimodal embedding \cite{drushchak-etal-2025-multimodal, Zhang_2025_CVPR, abootorabi2025askmodalitycomprehensivesurvey, Rackauckas_2024, mei2025surveymultimodalretrievalaugmentedgeneration}.

\begin{figure*}[t]
    \centering
    \includegraphics[width=.8\textwidth]{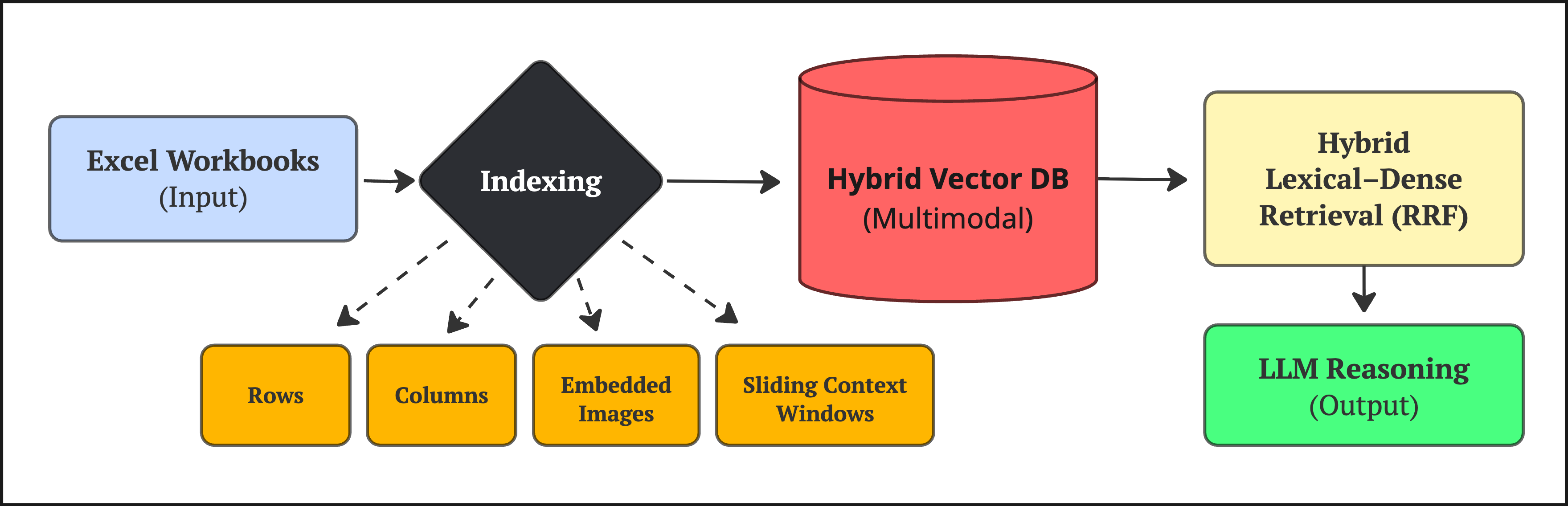}
    \Description{Architecture diagram showing the FRTR framework: Excel workbooks are decomposed into rows, columns, sliding context windows, and embedded images. These are indexed into a multimodal hybrid vector database. At query time, hybrid retrieval using RRF fetches relevant chunks which are passed to an LLM for reasoning.}
    \caption{Excel workbooks are decomposed into rows, columns, sliding context windows, and embedded images during indexing. These components are stored in a multimodal hybrid vector database supporting lexical and dense fields. Hybrid lexical–dense retrieval (RRF) fetches relevant evidence for LLM reasoning.}
    \label{fig:frtr_framework}
    
\end{figure*}

\subsection{Long Context, Retrieval, and Hybrid Search}

Despite advances in long-context modeling, evidence retrieval remains a bottleneck: simply extending context windows does not guarantee that models attend effectively to mid-range information \cite{liu2023lostmiddlelanguagemodels}. Retrieval-augmented generation approaches address this by selecting relevant evidence before generation, improving both performance and interpretability \cite{lewis2020rag, gao2024retrievalaugmentedgenerationlargelanguage}. Hybrid lexical-dense retrieval, fused through methods such as Reciprocal Rank Fusion (RRF) \cite{cormack2009rrf, azureHybridSearchRRF}, offers robustness and simplicity across tasks. However, basic chunking and retrieval methodologies do not take into account the complicated spatial relationships in spreadsheets, where logic can apply across rows, columns, and even sheets.

\subsection{Multimodal Spreadsheet Understanding and Document VQA}

Spreadsheets often convey semantics through visual layout, including bold headers, merged cells, and shaded totals, as well as through embedded images such as charts or scanned receipts. Vision–language models (VLMs) still struggle to interpret such spatial and structural cues \cite{radford2021clip, jia2021align, li2023blip2}. Parallel research in chart and document VQA (e.g., ChartQA \cite{masry2022chartqa}, DocVQA \cite{mathew2021docvqa}) highlights the challenges of numerically grounded visual reasoning. These methods typically treat visual and textual modalities in isolation, without integrating multimodal embeddings that unify text, layout, and image semantics.

\subsection{Benchmarks for Spreadsheet Interaction}

Beyond table QA, recent benchmarks target spreadsheet interaction directly. SpreadsheetLLM \cite{dong2025spreadsheetllmencodingspreadsheetslarge} performs reasoning on encoded spreadsheets, while SpreadsheetBench \cite{kim2024spreadsheetbench} introduces realistic, forum-based spreadsheet questions. InstructExcel \cite{zhang2024instructexcel} focuses on natural-language-to-OfficeScript translation. No existing benchmarks assess cross-sheet reasoning, multimodal queries, and large-scale evaluation together, reflecting the real-world challenges of enterprise workbooks.

\section{Framework Overview}
\label{sec:framework}

\subsection{Motivation and Design Goals}

Real-world Excel workbooks exhibit three intertwined challenges: scale, multimodality, and cross-sheet dependency. Workbooks can span thousands of rows across multiple worksheets, include embedded charts or receipts, and encode meaning through layout and formatting. Existing approaches either compress entire files into an LLM's context, trading precision for token cost, or rely on text-only embeddings that ignore visual cues. Although some enterprise spreadsheets contain irregular or weakly aligned layouts, FRTR does not assume well-formed tables; its use of overlapping row-, column-, and window-level units allows relevant evidence to be retrieved even when logical groupings are not strictly aligned with grid boundaries. 

FRTR addresses these challenges through three guiding principles:
\begin{enumerate}
    \item \textbf{Scalability}: Decompose workbooks into granular, retrievable units such as rows, columns, blocks, and images to bypass context limits, deduplicating merged cells and representing both formulas and calculated values.
    \item \textbf{Auditability}: Ensure transparent provenance via hybrid retrieval with RRF, supporting verifiable reasoning.
    \item \textbf{Multimodality}: Use joint text–vision embeddings so that both tabular and visual content contribute to retrieval and reasoning.
\end{enumerate}

\subsection{Architecture Overview}

As shown in Figure~\ref{fig:frtr_framework}, FRTR operates in three stages: \textit{Retrieve, Verify, and Compose}. The pipeline is formalized in Algorithm~\ref{alg:frtr}. During ingestion, Excel workbooks are decomposed into rows, columns, sliding windows, and embedded images (we investigate the impact of different chunk types in Section \ref{sec:ablation}). Each unit is serialized and embedded using a multimodal encoder (e.g., Amazon Titan Multimodal) and indexed in a hybrid vector database \cite{aws2025titanmultimodal}. At query time, FRTR retrieves the most relevant units using hybrid lexical–dense search and provides the fused results as evidence for LLM reasoning. This architecture reduces context size while preserving interpretability and scalability.

\subsection{Retrieval and Fusion Layer}

Each spreadsheet unit is indexed as a structured document with lexical, vector, and metadata fields. FRTR performs two parallel retrievals: BM25 for lexical matching and cosine similarity for dense embeddings, and fuses results via Reciprocal Rank Fusion (RRF) \cite{cormack2009rrf, robertson2009bm25}.
\[
\text{RRF}(d) = \sum_{r \in R} \frac{1}{k + \text{rank}_r(d)},
\]
where $k=60$ stabilizes low-ranked items. FRTR uses $K_v = K_s = 20$, returning the top-10 fused items annotated with provenance. This rank-based fusion avoids score normalization issues, ensuring robust, interpretable retrieval across modalities. RRF is used because it provides the most robust retrieval across both paraphrased queries and exact identifier lookups common in spreadsheet QA (see \ref{sec:ablation}).

\begin{algorithm}[t]
\small
\caption{FRTR: From Rows to Reasoning}
\label{alg:frtr}
\begin{algorithmic}[1]
\Require Workbook $\mathcal{W}$, query $q$, embedding model $\mathcal{E}$, LLM $\mathcal{M}$, hyperparameters $K_v, K_s, K, k{=}60$
\Ensure Answer with provenance

\State \textbf{// Stage 1: Indexing (Offline)}
\State $\mathcal{D} \leftarrow \emptyset$
\For{each sheet $S \in \mathcal{W}$}
    \State Compute window size: $s \leftarrow \lceil\sqrt{N / K_{\text{target}}}\rceil$
    \For{each unit $u \in \{\text{rows, columns, } s{\times}s \text{ windows, images}\}$}
        \State $\mathbf{v}_u \leftarrow \mathcal{E}(u)$ \Comment{Multimodal embedding}
        \State $\mathcal{D} \leftarrow \mathcal{D} \cup \{(\text{text}(u), \mathbf{v}_u, \text{metadata}(u))\}$
    \EndFor
\EndFor
\State Index $\mathcal{D}$ with lexical and dense fields

\State
\State \textbf{// Stage 2: Hybrid Retrieval (Online)}
\State $\mathbf{q} \leftarrow \mathcal{E}_{\text{text}}(q)$
\State $\mathcal{R}_v \leftarrow$ top-$K_v$ by $\cos(\mathbf{q}, \mathbf{v}_d)$ \Comment{Dense search}
\State $\mathcal{R}_s \leftarrow$ top-$K_s$ by BM25$(q, \text{text}_d)$ \Comment{Lexical search}
\For{$d \in \mathcal{R}_v \cup \mathcal{R}_s$}
    \State $\text{RRF}(d) \leftarrow \sum_{r \in \{v, s\}} \frac{1}{k + \text{rank}_r(d)}$
\EndFor
\State $\mathcal{C} \leftarrow$ top-$K$ by RRF score with provenance labels

\State
\State \textbf{// Stage 3: Answer Composition}
\State Construct prompt: $\mathcal{P} \leftarrow \{q, \mathcal{C}, \text{instruction}\}$
\State $\text{response} \leftarrow \mathcal{M}(\mathcal{P})$
\State \Return Parse JSON: $\{\text{reasoning}, \text{answer}\}$
\end{algorithmic}
\end{algorithm}

\subsection{Multimodal Embedding Layer}

For text-based units, FRTR serializes contextual information such as headers and indices, preserving spatial cues:
\begin{itemize}
    \item Rows: include column headers.
    \item Columns: include row indices.
    \item Windows: preserve $s \times s$ spatial layout.
\end{itemize}
These are embedded via the text encoder. Image units such as charts, receipts, and diagrams are embedded using the vision branch of the same model, ensuring a shared latent space. In our implementation, we encode images as base64 and pass them to the Titan Image-v1 embeddings API to generate embeddings. This allows queries such as "Q4 revenue trends" to retrieve both numerical data and corresponding charts. Generated embeddings $\mathbf{v}_u$ are stored in a unified vector field that holds both text and image representations. This unified field enables cross-modal retrieval within a single index, maintaining simplicity and generality.

\subsection{Reasoning and Output Generation}

After retrieval, FRTR invokes an LLM (e.g., GPT-5 or Claude Sonnet 4.5) with a structured prompt containing the query, retrieved chunks, and reasoning instructions. For chunks containing images, those images are passed into the LLM as attachments. Each chunk is formatted with explicit metadata:
\begin{verbatim}
Chunk 1 (Score: 0.0164, Source: Vector)
Type: row | Sheet: Sales_Q4
ROW_42: Product | Units | Revenue | ...
\end{verbatim}
The LLM synthesizes reasoning across these snippets, producing structured JSON outputs:
\begin{verbatim}
{
  "reasoning": "Analyzed columns B and C for revenue values.",
  "answer": "SUM(B2:B15)"
}
\end{verbatim}
This format ensures both interpretability and machine-readability, facilitating downstream automation and audit compliance. The complete prompt template used in this study is provided in Appendix~\ref{app:prompt}. FRTR itself does not execute formulas or modify spreadsheets, maintaining human oversight while ensuring verifiable, grounded reasoning.

\section{FRTR-Bench: Benchmark for Spreadsheet Reasoning}
\label{sec:benchmark}

\subsection{Dataset Construction}

Existing benchmarks like SpreadsheetLLM \cite{dong2025spreadsheetllmencodingspreadsheetslarge} and SpreadsheetBench \cite{kim2024spreadsheetbench} focus on single-sheet, text-only reasoning. To evaluate cross-sheet and multimodal reasoning at scale, we introduce \textbf{FRTR-Bench}, a dataset of 30 multimodal Excel workbooks across diverse domains. Similar to how Jackal introduced an execution-based, real-world benchmark for text-to-JQL tasks to evaluate LLMs under production-like conditions, we design FRTR-Bench to capture enterprise spreadsheet reasoning at realistic scale and modality \cite{frank2025jackal}. Each workbook stresses three dimensions: 
(1) Scale: up to 210,000 rows per workbook, far beyond typical LLM context; 
(2) Multimodality: embedded PNG images such as charts, receipts, and dashboards; and 
(3) Cross-sheet complexity: multi-sheet references (e.g., \texttt{=SUM(Sheet1!B2:B100, Sheet2!C5:C50)}).  

Workbooks are divided into difficulty tiers by scale:
\begin{itemize}
    \item \textbf{Easy}: $<5{,}000$ rows - small business summaries.
    \item \textbf{Medium}: $5{,}000$--$20{,}000$ rows - mid-size audit or operations data.
    \item \textbf{Hard}: $20{,}000$--$210{,}000$ rows - large-scale consolidation workbooks.
\end{itemize}

Domains include (but are not limited to) finance, supply chain, healthcare, energy, government, education, and others. Each workbook includes a metadata sheet, 1–5 data sheets, embedded PNGs with captions, and a \textit{Questions} sheet containing 5–10 manually crafted queries with explicit provenance (e.g., \texttt{Sheet1!B5} or \texttt{Image ID: Chart\_001}).

\subsection{Evaluation Protocol}

We evaluate FRTR-Bench on two axes: (1) answer accuracy, (2) efficiency metrics capturing latency and prompt length.

\textbf{Answer Accuracy.} Computed by comparing model outputs with annotated ground-truth answers and provenance (cells, formulas, or images). Answers are considered correct if numerically consistent or functionally identical in formula logic (e.g., \texttt{SUM(B2:B10)} vs.\ \texttt{SUM(B2:B5)+SUM(B6:B10)}).

\textbf{Latency.} Average response time per query (not including embedding and retrieval), measured from user query submission to model output generation. This metric reflects generation overhead.

\textbf{Mean Tokens.} Mean number of tokens per model input after retrieval and formatting. Lower token counts indicate more efficient retrieval and reduced context consumption.

\subsection{Benchmark Statistics and Examples}

\begin{table}[t]
\centering
\caption{FRTR-Bench dataset statistics summarizing workbook, sheet, and multimodal content scale.}
\label{tab:frtr_stats}
\begin{tabular}{lr}
\toprule
\textbf{Metric} & \textbf{Count} \\
\midrule
Workbooks & 30 \\
Sheets & 155 \\
Rows & 656,457 \\
Cells & 3,928,934 \\
Embedded Images & 53 \\
Cross-Sheet Formulas & 30 \\
Questions & 157 \\
\bottomrule
\end{tabular}
\end{table}

As shown in Table~\ref{tab:frtr_stats}, FRTR-Bench contains 30 Excel workbooks (155 sheets, 53 images, 656k rows, and approximately 3.9 million cells), each including embedded images and cross-sheet formulas. The benchmark captures realistic enterprise workflows with diverse natural-language queries that require lookup, aggregation, and cross-sheet reasoning.

For example, in \texttt{frtr\_0003\_consolidation.xlsx} (Hard, 49,867 rows):
\begin{quote}
\small
\noindent\textbf{Question:} What is the consolidated operating income for Region EMEA in FY2024?\\
\textbf{ReasoningType:} cross-sheet\\
\textbf{Provenance:} \texttt{Consolidation!E47},\\
\hspace*{1.5em}\texttt{=SUM(EMEA\_Sales!D:D,EMEA\_Ops!E:E)}\\
\textbf{Answer:} \$12{,}450{,}000
\end{quote}
\normalsize

\begin{table*}[t]
\centering
\caption{Performance comparison on FRTR-Bench. FRTR consistently outperforms SpreadsheetLLM while using 40\% fewer tokens. Mean tokens: FRTR 7,691; SpreadsheetLLM 12,745. We implemented the SpreadsheetLLM algorithm as written \cite{spreadsheetllm_github}.}
\label{tab:exp1_results}
\setlength{\tabcolsep}{10pt}
\renewcommand{\arraystretch}{1.1}
\small
\begin{tabular}{l|cc|cc|c}
\toprule
 & \multicolumn{2}{c|}{\textbf{FRTR (Ours)}} & \multicolumn{2}{c|}{\textbf{SpreadsheetLLM}} & \\
\textbf{Model} & \textbf{Accuracy} & \textbf{Latency (s)} & \textbf{Accuracy} & \textbf{Latency (s)} & \textbf{$\Delta$ Accuracy} \\
\midrule
GPT-4o & 0.49 & 5.04 & 0.06 & 1.14 & \textbf{+0.43} \\
GPT-5 & \underline{0.73} & 15.50 & 0.18 & 18.10 & \textbf{+0.55} \\
Claude Sonnet 4.5 & \textbf{0.74} & 11.71 & 0.13 & 8.80 & \textbf{+0.61} \\
Gemini 2.5 Pro & 0.67 & 26.90 & 0.24 & 33.35 & \textbf{+0.43} \\
LLaMA Maverick-17B & 0.56 & 2.42 & 0.23 & 1.15 & \textbf{+0.33} \\
GPT-OSS-120B & 0.51 & 5.75 & 0.21 & 8.80 & \textbf{+0.30} \\
\bottomrule
\end{tabular}
\end{table*}

A full-context LLM would need to serialize hundreds of thousands of cells, far beyond feasible context limits, while FRTR retrieves fewer than 10 relevant chunks ($<$5K tokens) to reason effectively. FRTR-Bench thus provides a rigorous, multimodal, and scalable benchmark for retrieval-augmented spreadsheet reasoning, capturing the challenges of real-world analytical workflows.

\section{Evaluation}
\label{sec:evaluation}

This section presents two complementary experiments evaluating the proposed FRTR framework across diverse settings. The first experiment assesses scalability and multimodal reasoning using the newly introduced FRTR-Bench, while the second evaluates generalization and efficiency on the established SpreadsheetLLM benchmark. Together, they demonstrate that retrieval-first multimodal reasoning consistently outperforms compression- or context-based methods.

\paragraph{Metrics.}
All experiments use the same evaluation criteria defined in Section~\ref{sec:benchmark}: 
Answer Accuracy, Latency (s), and Mean Tokens. Latency refers to the inference time of the LLM.

\paragraph{LLMs.}
All experiments evaluate six LLMs spanning both small and large, and open- and closed-weight architectures: GPT-4o, GPT-5, Claude Sonnet 4.5, Gemini 2.5 Pro, LLaMA Maverick-17B Instruct, and GPT-OSS-120b. All LLMs and embedding models were accessed via APIs using standard pricing. While FRTR incurs additional embedding calls during indexing, embedding models are substantially cheaper than LLM inference, so the overall cost is dominated by prompt tokens passed to the LLM and remains lower than full-context baselines.

\paragraph{Baselines.}
We compare FRTR against two baseline approaches:
\begin{enumerate}
    \item \textbf{SpreadsheetLLM} – state-of-the-art configuration using SheetEncoder and SheetCompressor.
    \item \textbf{Naïve Long-Context} – direct serialization of entire sheets into the prompt, limited only by model context length.
\end{enumerate}

\begin{figure*}[!t]
    \centering
    \includegraphics[width=0.8\textwidth]{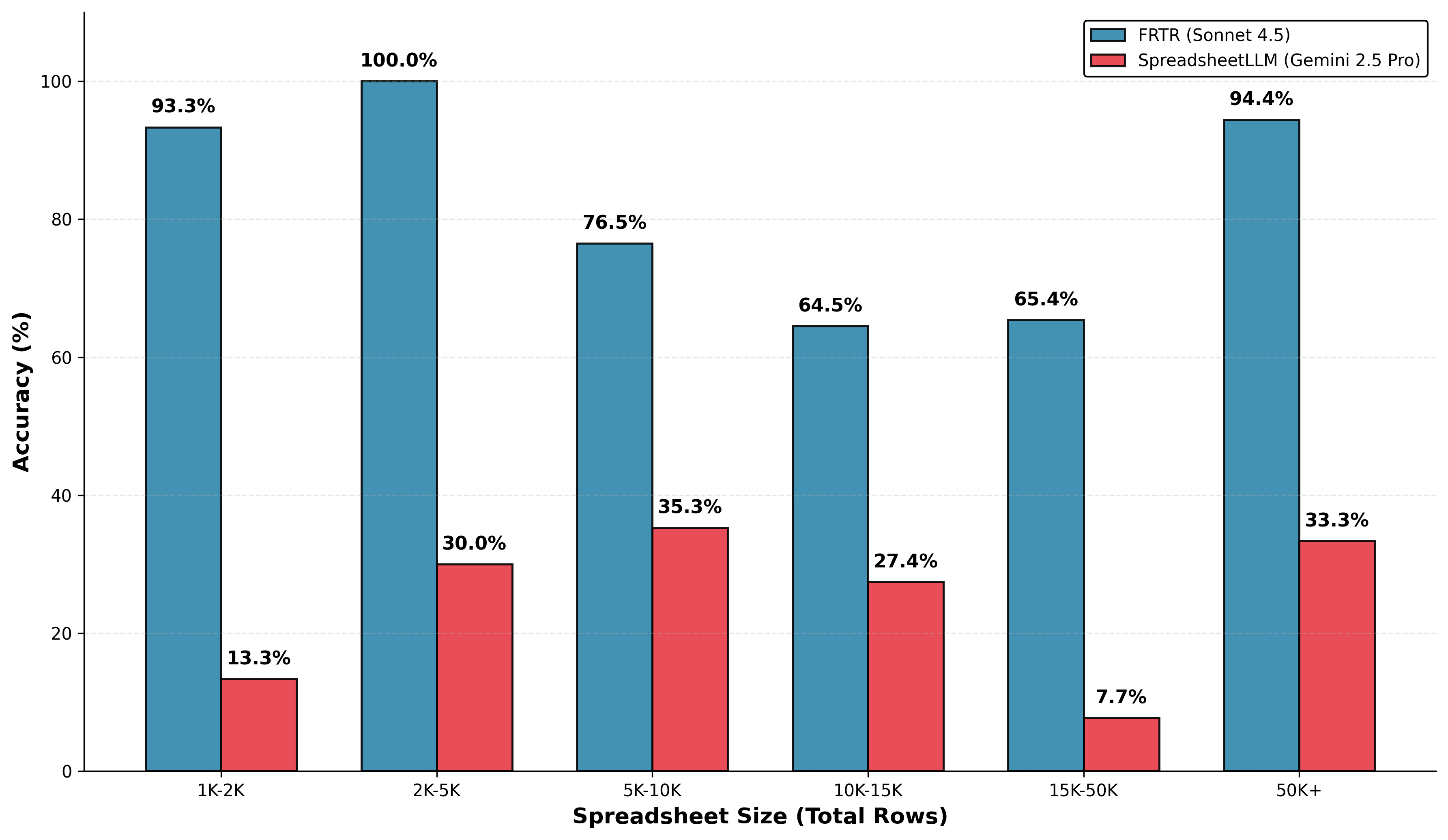}
    \Description{Line chart comparing peak accuracy of FRTR and SpreadsheetLLM methods across spreadsheets with varying total row counts. FRTR consistently outperforms SpreadsheetLLM, with the gap widening as spreadsheet size increases.}
    \caption{Peak performance comparison by total rows across methods on FRTR-Bench.}
    \label{fig:performance_by_difficulty}
\end{figure*}

\subsection{Experiment 1 - FRTR-Bench Evaluation}
\label{sec:exp1}

\subsubsection{Goal}
The first experiment evaluates FRTR on the newly introduced FRTR-Bench dataset, assessing its ability to handle large, multimodal, and cross-sheet enterprise workbooks. We compare FRTR to SpreadsheetLLM, the strongest prior single-sheet compression baseline, across six large language models (LLMs) and three difficulty tiers (\textit{Easy}, \textit{Medium}, \textit{Hard}). The objective is to measure reasoning accuracy, token efficiency, and latency in realistic spreadsheet settings that include multimodal content and multi-sheet dependencies. Since SpreadsheetLLM was designed for single-sheet reasoning, we extended it to multi-sheet settings by compressing each sheet individually and concatenating all compressed sheets into a single prompt. This concatenation fit within model context limits without truncation, representing the most favorable adaptation of their approach to cross-sheet tasks.

\subsubsection{Experimental Setup}
Both systems were tested on identical workbooks and queries drawn from FRTR-Bench. Each query was evaluated for correctness, reasoning trace, and provenance alignment. FRTR employed hybrid lexical–dense retrieval with RRF using the Amazon Titan Multimodal encoder, while SpreadsheetLLM followed its published compression–encoding approach \cite{dong2025spreadsheetllmencodingspreadsheetslarge}. All experiments were executed under identical prompt templates and model configurations.

\subsubsection{Results Analysis}
Table~\ref{tab:exp1_results} summarizes the comparative performance of FRTR and SpreadsheetLLM on FRTR-Bench. Across all difficulty levels, FRTR consistently outperforms SpreadsheetLLM. On easy workbooks, FRTR attains mean accuracies between 0.86 and 0.93 for top-tier models, exceeding SpreadsheetLLM's 0.18–0.23 by more than two times. Performance remains robust on medium (0.62–0.65) and hard (0.66–0.72) tiers, where SpreadsheetLLM collapses below 0.10. Figure~\ref{fig:performance_by_difficulty} displays the average results by difficulty across both FRTR and SpreadsheetLLM. Comprehensive results for each model by difficulty mode are reported in Appendix~\ref{app:results}. This confirms that compression-based methods do not scale when cross-sheet references and multimodal evidence are required, while the FRTR retrieval-based design preserves contextual fidelity at scale. FRTR also demonstrates high token efficiency: the average prompt length is only 7.7K tokens, compared to SpreadsheetLLM's 13.1K tokens, corresponding to a mean compression ratio of 251× relative to raw workbook size. Despite smaller prompts, FRTR achieves higher accuracy, indicating that selective retrieval retains the most relevant evidence.

\subsubsection{Discussion}
The FRTR-Bench evaluation demonstrates that retrieval-first architectures outperform context-compression systems for large, multimodal spreadsheets. FRTR consistently achieves 2–4× higher accuracy, 40\% lower token usage, and comparable or lower latency than SpreadsheetLLM, while generalizing across diverse LLM families. Figure~\ref{fig:performance_by_difficulty} demonstrates FRTR's high accuracy across various spreadsheet lengths. The U-shaped accuracy pattern suggests that structural complexity, not size alone, drives difficulty. Very large spreadsheets in the benchmark are primarily transaction logs with repetitive structure and simple aggregation queries (e.g., payroll records, time-series data), enabling effective retrieval and reasoning. In contrast, medium spreadsheets often represent complex business scenarios (consolidations, multi-entity reporting) with multiple interconnected worksheets, cross-references, and semantically similar sections that create retrieval ambiguity. This 'confusion zone' challenges both approaches: FRTR's retrieval must disambiguate between multiple similar contexts, while SpreadsheetLLM's compression discards critical cross-sheet relationships.

Notably, FRTR achieves 65\% accuracy on the 23 image-dependent questions in FRTR-Bench, enabling a class of multimodal queries that text-only approaches cannot address. Once the spreadsheet was indexed and embedded, retrieval time was negligible and inference time (as reported in Table~\ref{tab:exp1_results}) was relatively consistent across approaches. These results validate FRTR-Bench as a challenging benchmark and establish FRTR as a state-of-the-art framework for advanced multimodal, retrieval-augmented spreadsheet understanding.

\subsection{Experiment 2 - SpreadsheetLLM Benchmark Evaluation}
\label{sec:exp2}

\subsubsection{Goal}
The second experiment evaluates whether FRTR can match or approach the accuracy of the state-of-the-art SpreadsheetLLM while offering superior interpretability and efficiency. We compare FRTR against SpreadsheetLLM's compression-based framework and a naïve long-context baseline to assess whether retrieval-based reasoning can achieve comparable performance with significantly lower token usage and improved transparency.

\subsubsection{Experimental Settings}
We use the SpreadsheetLLM benchmark \cite{dong2025spreadsheetllmencodingspreadsheetslarge}, which consists of single-sheet, text-only question–answer pairs requiring numerical reasoning, formula synthesis, and contextual lookup comprehension. Unlike FRTR-Bench, this dataset excludes multimodal and cross-sheet reasoning, emphasizing tabular text understanding within context-limited settings. For SpreadsheetLLM's evaluation on this benchmark, the compressed oracle sheet is provided as input, following the methodology detailed in SpreadsheetLLM \cite{dong2025spreadsheetllmencodingspreadsheetslarge}.

\begin{table}[t]
  \centering
  \caption{Performance comparison on the SpreadsheetLLM benchmark. FRTR achieves answer accuracy comparable to SpreadsheetLLM. Mean tokens: FRTR 6,920; SpreadsheetLLM 5,811; Na\"ive 13,631.}
  \label{tab:exp2_results}
  \setlength{\tabcolsep}{4pt}
  \renewcommand{\arraystretch}{1.05}
  \footnotesize
  
  \begin{tabular}{@{}l
                  >{\raggedright\arraybackslash}p{0.40\columnwidth}
                  cc@{}}
  \toprule
  \textbf{Technique} & \textbf{Model} &
  \makecell{\textbf{Answer}\\\textbf{Accuracy}} &
  \makecell{\textbf{Latency}\\\textbf{(s)}} \\
  \midrule
  \textbf{FRTR} & GPT-4o & 0.77 & 2.48 \\
   & GPT-5 & 0.87 & 18.06 \\
   & Claude Sonnet 4.5 & 0.84 & 10.40 \\
   & Gemini 2.5 Pro & 0.85 & 24.06 \\
   & LLaMA Maverick-17B-instruct & 0.53 & 2.18 \\
   & GPT-OSS-120b & 0.75 & 7.37 \\
  \midrule
  \textbf{SpreadsheetLLM} & GPT-4o & 0.78 & 0.81 \\
   & GPT-5 & 0.90 & 5.50 \\
   & Claude Sonnet 4.5 & 0.91 & 5.49 \\
   & Gemini 2.5 Pro & 0.89 & 10.00 \\
   & LLaMA Maverick-17B-instruct & 0.73 & 0.73 \\
   & GPT-OSS-120b & 0.80 & 4.01 \\
  \midrule
  \textbf{\makecell[l]{Na\"ive\\Full-Context}} & GPT-4o & 0.35 & 3.01 \\
   & GPT-5 & 0.68 & 18.07 \\
   & Claude Sonnet 4.5 & 0.55 & 9.06 \\
   & Gemini 2.5 Pro & 0.67 & 15.48 \\
   & LLaMA Maverick-17B-instruct & 0.41 & 2.45 \\
   & GPT-OSS-120b & 0.59 & 8.61 \\
  \bottomrule
  \end{tabular}
  \end{table}


\begin{figure*}[t]
\centering
\includegraphics[width=0.83\textwidth]{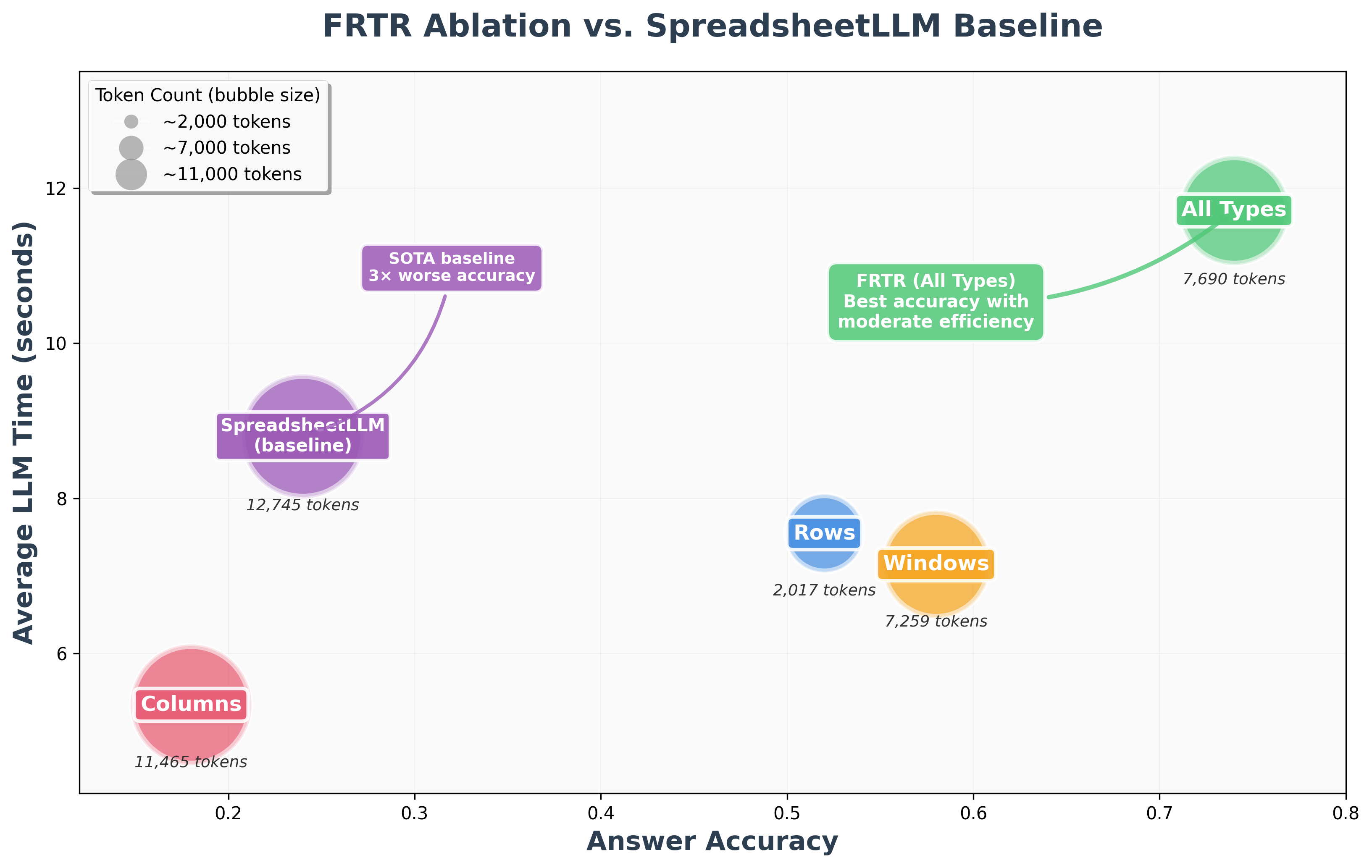}
\caption{Unit granularity ablation: accuracy vs. latency, bubble size = token count (Claude Sonnet 4.5, FRTR-Bench). \textit{All Types} (FRTR) achieves 0.74 accuracy with 7,690 tokens, outperforming SpreadsheetLLM with Gemini 2.5 Pro (0.24, 12,745 tokens).}
\label{fig:unit_ablation_bubble}
\end{figure*}

\subsubsection{Results Analysis}
As shown in Table~\ref{tab:exp2_results}, FRTR performs on par with SpreadsheetLLM in overall accuracy across most model families while substantially outperforming the naïve full-context baseline. For high-capacity models such as \texttt{GPT-5}, \texttt{Claude Sonnet 4.5}, and \texttt{Gemini 2.5 Pro}, FRTR attains accuracies in the 0.84–0.87 range, closely matching SpreadsheetLLM's 0.89–0.91. Open-weight models follow the same trend, achieving strong relative performance despite having smaller token counts.  
Importantly, FRTR achieves this parity while using roughly 40–50\% fewer tokens than the full-context baseline and maintaining stable latency across models. The retrieval-based design also provides explicit provenance for every reasoning step, enabling interpretability that compression-based systems lack.

\subsubsection{Discussion}
Although FRTR does not surpass SpreadsheetLLM in raw accuracy on this single-sheet benchmark, it achieves near-equivalent results while offering stronger efficiency and transparency benefits. It is important to further note that FRTR's approach considered every sheet in an Excel workbook, while SpreadsheetLLM was given the oracle sheet (as was implemented in the SpreadsheetLLM paper). FRTR's advanced multimodal retrieval-augmented approach focuses on retrieving the minimal evidence needed for reasoning rather than compressing the entire sheet, reducing token consumption and supporting explainable, auditable outputs.  
These findings indicate that FRTR generalizes effectively even in non-multimodal, single-sheet scenarios where its architectural advantages (cross-sheet retrieval and image reasoning) are not fully utilized. Matching SpreadsheetLLM's performance under these constrained conditions highlights the robustness of retrieval-augmented reasoning and its potential as a scalable, interpretable alternative to context-compression approaches.

\subsection{Ablation Studies}
\label{sec:ablation}

To validate the design choices in FRTR, we conduct ablation experiments on FRTR-Bench using the highest performing LLM (Claude Sonnet 4.5), examining two key dimensions: retrieval methodology and unit granularity. In both cases, we retrieve the top 10 most relevant documents.

\subsubsection{Retrieval Method Ablation}
We compare three retrieval strategies on FRTR-Bench to assess the contribution of hybrid search: (1) Hybrid RRF combining lexical and dense retrieval, (2) Vector-only dense retrieval, and (3) BM25 lexical-only retrieval.

\begin{table}[h]
\centering
\caption{Retrieval method ablation on FRTR-Bench. Results show answer accuracy, retrieval time, and tokens retrieved across different search strategies.}
\label{tab:retrieval_ablation}
\setlength{\tabcolsep}{6pt}
\begin{tabular}{lccc}
\toprule
\textbf{Retrieval Method} & \textbf{\shortstack{Answer\\Accuracy}} & \textbf{\shortstack{Avg Retrieval\\Time (s)}} & \textbf{\shortstack{Avg\\Tokens}} \\
\midrule
Hybrid (RRF) & \textbf{0.74} & 0.82 & 6,903 \\
Vector Only & 0.59 & 0.70 & 5,857 \\
BM25/Lexical Only & 0.42 & \textbf{0.11} & \textbf{5,616} \\
\bottomrule
\end{tabular}
\end{table}

The results demonstrate that hybrid RRF retrieval achieves significantly higher accuracy (0.74) compared to vector-only (0.59) and lexical-only (0.42) approaches. While BM25 offers the fastest retrieval time (0.11s), its accuracy is insufficient for complex spreadsheet reasoning. Vector-only retrieval improves upon lexical search but still falls short of hybrid performance, suggesting that semantic similarity alone misses exact-match patterns captured by BM25. The hybrid approach combines the strengths of both methods through rank fusion, achieving the best accuracy with only a modest increase in retrieval time (0.82s). This validates FRTR's design choice to employ RRF-based hybrid retrieval for robust, scalable spreadsheet understanding.

\subsubsection{Unit Granularity Ablation}
We evaluate different indexing granularities by comparing retrieval using only rows, columns, or windows against our default configuration, which combines all unit types. In all cases, images are included in the retrieval set. This ablation isolates the contribution of each granularity level to overall reasoning performance.

\begin{table}[t]
\centering
\caption{Unit granularity ablation on FRTR-Bench. Results demonstrate the complementary value of multi-granular indexing for spreadsheet reasoning.}
\label{tab:unit_ablation}
\setlength{\tabcolsep}{6pt}
\begin{tabular}{lccc}
\toprule
\textbf{Units Retrieved} & \textbf{\shortstack{Answer\\Accuracy}} & \textbf{\shortstack{Avg LLM\\Time (s)}} & \textbf{\shortstack{Avg\\Tokens}} \\
\midrule
Rows & 0.52 & 7.55 & 2,017 \\
Columns & 0.18 & 5.34 & 11,465 \\
Windows & 0.58 & 7.15 & 7,259 \\
All Types & \textbf{0.74} & 11.71 & 7,690 \\
\bottomrule
\end{tabular}
\end{table}

The results reveal that different granularities capture complementary information, as visualized in Figure~\ref{fig:unit_ablation_bubble}. Row-only retrieval achieves 0.52 accuracy, insufficient for cross-column reasoning. Column-only retrieval performs poorly (0.18 accuracy), as column serialization lacks sufficient context and produces verbose representations. Window-based retrieval achieves 0.58 accuracy, capturing spatial relationships within localized regions.

Combining all unit types (rows + columns + windows + images) achieves the highest accuracy (0.74), demonstrating that multi-granular indexing surfaces relevant evidence regardless of query structure. Notably, FRTR's multi-granular approach substantially outperforms the SpreadsheetLLM baseline (0.24 accuracy), achieving over 3× improvement while maintaining comparable token efficiency. While this incurs slightly higher latency (11.71s) due to increased retrieval and reasoning complexity, the accuracy gain justifies the cost. Token cost variation is due to benchmark characteristics: most spreadsheets had long tables with few columns, resulting in longer column documents and shorter row documents. This confirms that FRTR's multi-granular design is essential for robust spreadsheet reasoning across diverse query types. 

\subsection{Cost Analysis}
LLM inference costs scale with token consumption. Table~\ref{tab:exp1_results} shows FRTR uses 7,691 tokens on average vs. SpreadsheetLLM's 12,745 tokens, a 40\% reduction. At typical API rates (\$0.01–0.03 per 1K input tokens for frontier models), this translates to significant savings at scale. For an enterprise processing 1,000 queries daily, FRTR would reduce annual LLM costs by approximately 40\%, while improving accuracy. Embedding costs are incurred once during indexing and amortized across all subsequent queries.

\section{Limitations}
While FRTR demonstrates strong performance, certain design choices merit discussion. First, our multimodal embedding approach achieves 65\% accuracy on image-dependent questions, a capability that would otherwise be zero without multimodal retrieval, and we anticipate further gains as enterprise-grade multimodal encoders continue to mature. Second, FRTR-Bench focuses on English-language spreadsheets, though the retrieval architecture is language-agnostic and extensible to multilingual settings. Third, FRTR generates formulas and cell references rather than executing them directly, preserving human oversight aligned with enterprise audit requirements. Finally, while retrieval latency is negligible once indexed (under 1 second), initial workbook indexing is a one-time cost amortized across all subsequent queries. The initially substantial indexing time was alleviated by employing batch processing and parallelization techniques.

\section{Ethics and Reproducibility}
FRTR-Bench consists of synthetic enterprise-style workbooks designed to simulate realistic analytical scenarios without using real user data or proprietary information. The benchmark is available via a github repository\footnote{\href{https://github.com/AnmolGulati6/FRTR-bench}{ FRTR-Bench Repository (https://github.com/AnmolGulati6/FRTR-bench)}}.

\section{Conclusion}
Large Language Models (LLMs) continue to struggle with reasoning over large-scale, enterprise-grade spreadsheets that contain thousands of rows, multiple linked sheets, and multimodal elements such as charts and receipts. Existing approaches based on single-sheet compression or full-context encoding fail to scale and do not reflect how real users interact with complex analytical workbooks. To address these challenges, we introduced From Rows to Reasoning (FRTR), an advanced, multimodal retrieval-augmented generation framework that decomposes spreadsheets into granular row, column, and block embeddings, employs hybrid lexical–dense retrieval with Reciprocal Rank Fusion (RRF), and integrates multimodal embeddings to reason jointly over numerical and visual content. Alongside this framework, FRTR-Bench provides the first large-scale benchmark for multimodal spreadsheet reasoning, comprising 30 enterprise-grade Excel workbooks spanning nearly four million cells and more than 50 embedded images. We evaluated FRTR using six LLMs, achieving 74\% answer accuracy on FRTR-Bench, achieving 50 percentage points higher accuracy than prior state-of-the-art approaches (74\% vs. 24\%). On the SpreadsheetLLM benchmark, FRTR achieved 87\% accuracy while reducing token usage by roughly 50\% compared to a full-context baseline. These results show that retrieval-first architectures enable scalable, interpretable, and efficient spreadsheet reasoning. FRTR gives LLMs the critical ability to reason over massive spreadsheets in domains such as finance, auditing, and healthcare, where verifiability and efficiency are critical. Future work should explore adaptive retrieval, finer multimodal alignment, and agentic spreadsheet manipulation.

\clearpage

\bibliographystyle{ACM-Reference-Format}
\bibliography{references}

\clearpage
\appendix
\section{Experiment 1 - FRTR-Bench Evaluation}
\label{app:results}

\begin{strip}
\centering
\scriptsize
\renewcommand{\arraystretch}{1.05}
\setlength{\tabcolsep}{4pt}
\captionof{table}{Detailed per-model results on FRTR-Bench grouped by difficulty tier. Times are in seconds; tokens are prompt tokens.}
\label{tab:appendix_combined_results}
\begin{adjustbox}{center,max width=\textwidth}
\begin{tabular}{llcccccc}
\toprule
\textbf{Difficulty} & \textbf{Model} &
\multicolumn{3}{c}{\textbf{FRTR}} &
\multicolumn{3}{c}{\textbf{SpreadsheetLLM}} \\
\cmidrule(lr){3-5} \cmidrule(lr){6-8}
 & & \makecell{\textbf{Accuracy}} & \makecell{\textbf{Tokens}} & \makecell{\textbf{Latency}\\(s)} &
   \makecell{\textbf{Accuracy}} & \makecell{\textbf{Tokens}} & \makecell{\textbf{Latency}\\(s)} \\
\midrule

\textbf{Easy} 
 & Gemini 2.5 Pro & \underline{0.864} & 12{,}088 & 21.38 & 0.205 & 6{,}673 & 20.97 \\
 & GPT-5 & \underline{0.864} & 12{,}088 & 13.29 & 0.227 & 6{,}673 & 14.60 \\
 & Claude Sonnet 4.5 & \textbf{0.932} & 12{,}088 & 10.79 & 0.182 & 6{,}673 & 7.12 \\
 & GPT-4o & 0.545 & 12{,}088 & 3.97 & 0.091 & 6{,}673 & 0.89 \\
 & LLaMA Maverick-17B & 0.750 & 12{,}088 & 2.25 & 0.205 & 6{,}673 & 0.90 \\
 & GPT-OSS-120b & 0.545 & 12{,}088 & 5.00 & 0.000 & 6{,}673 & 7.68 \\
\midrule

\textbf{Medium}
 & Gemini 2.5 Pro & 0.557 & 7{,}235 & 30.84 & 0.278 & 17{,}483 & 38.85 \\
 & GPT-5 & \underline{0.646} & 7{,}235 & 16.10 & 0.114 & 17{,}483 & 20.68 \\
 & Claude Sonnet 4.5 & \textbf{0.658} & 7{,}235 & 12.31 & 0.089 & 17{,}483 & 9.68 \\
 & GPT-4o & 0.443 & 7{,}235 & 5.08 & 0.101 & 17{,}483 & 1.33 \\
 & LLaMA Maverick-17B & 0.443 & 7{,}235 & 2.44 & 0.245 & 17{,}483 & 1.46 \\
 & GPT-OSS-120b & 0.468 & 7{,}235 & 6.59 & 0.148 & 17{,}483 & 9.93 \\
\midrule

\textbf{Hard}
 & Gemini 2.5 Pro & \underline{0.694} & 3{,}318 & 24.99 & 0.194 & 9{,}173 & 35.98 \\
 & GPT-5 & \textbf{0.722} & 3{,}318 & 16.88 & 0.250 & 9{,}173 & 16.70 \\
 & Claude Sonnet 4.5 & 0.667 & 3{,}318 & 11.53 & 0.167 & 9{,}173 & 8.75 \\
 & GPT-4o & 0.528 & 3{,}318 & 5.17 & 0.083 & 9{,}173 & 1.02 \\
 & LLaMA Maverick-17B & 0.583 & 3{,}318 & 2.25 & 0.222 & 9{,}173 & 0.78 \\
 & GPT-OSS-120b & 0.556 & 3{,}318 & 4.82 & 0.083 & 9{,}173 & 7.69 \\
\bottomrule
\end{tabular}
\end{adjustbox}
\normalsize
\end{strip}

\section{Token Usage Visualization}
\label{app:tokenusage}

\begin{figure}[H]
\centering
\includegraphics[width=\textwidth]{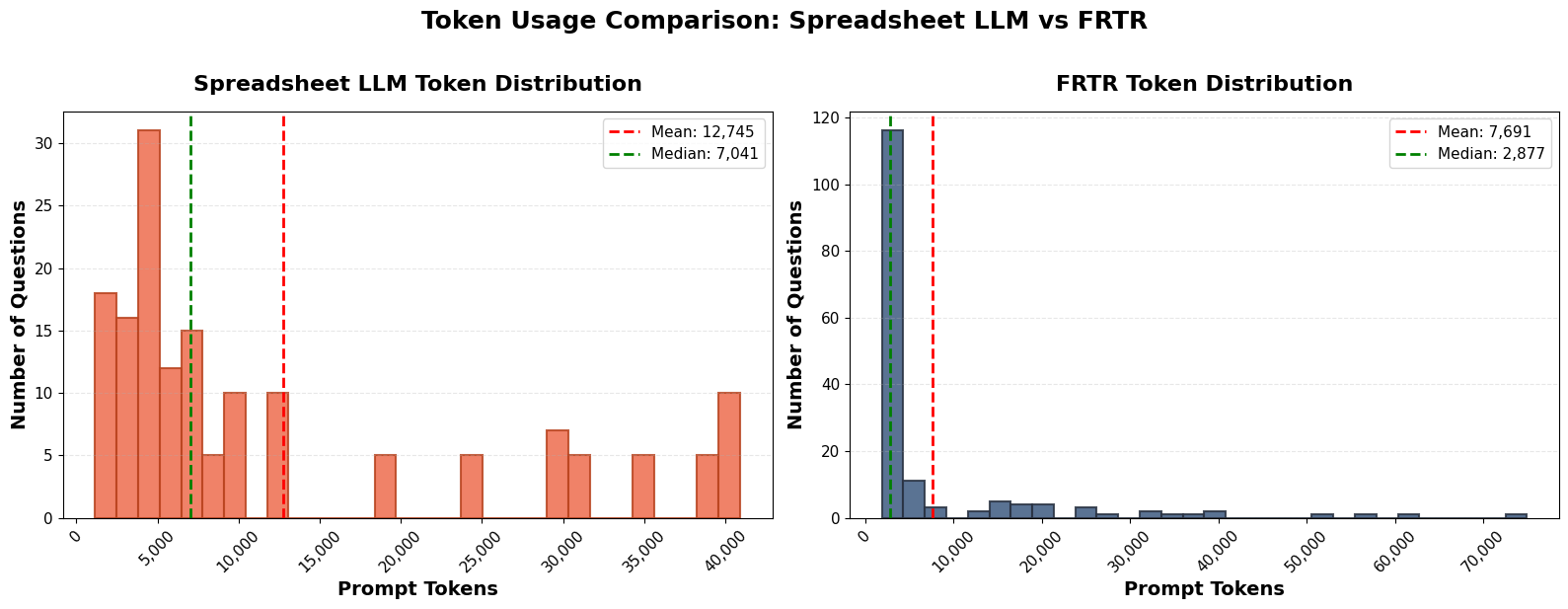}
\Description{Bar chart comparing prompt token usage between FRTR and SpreadsheetLLM across different spreadsheet files. FRTR uses significantly fewer tokens than SpreadsheetLLM due to its retrieval-based approach.}
\caption{Visualization of prompt token usage for FRTR and SpreadsheetLLM.}
\label{fig:tokenusage}
\end{figure}

\clearpage
\section{Prompt Template}
\label{app:prompt}

\begin{center}
\begin{minipage}{\textwidth}
\begin{lstlisting}[style=prompt,
  linewidth=\textwidth,
  xleftmargin=0pt,
  framexleftmargin=0pt
]
You are an Excel formula generator with reasoning capabilities. Based on the question and spreadsheet data provided, you will return the cell reference or formula AND explain your reasoning in JSON format.

You will be provided four potential types of chunks: columns, rows, windows of cells, and embedded images. Just because a chunk is listed below does not mean it contains the answer.

You will be provided relevant chunks from the spreadsheet. Keep in mind that headers and values are not separated, particularly for column chunks, so the values you receive could be part of the headers OR part of the actual table.

Headers could persist beyond the first row, so be careful when interpreting the data.

Instructions:
- Analyze the provided chunks and determine the answer
- Explain which chunks are relevant and why
- Describe your thought process for arriving at the answer
- Return your response as a valid JSON object with two keys: "reasoning" and "answer"

**CRITICAL RULES FOR IMAGES/CHARTS:**
- **IF the question asks about a TREND, PATTERN, or VISUAL INSIGHT from an image/chart: DESCRIBE IT IN WORDS, NOT cell references**
- **IF an image shows a chart/graph and the question asks "what is the trend?", "what pattern?", "what does it show?": Answer with descriptive text like "increasing trend", "declining over time", "peaked in Q3", etc.**
- **NEVER answer trend/pattern questions with cell references like "A5" or "Sheet1!B2"**
- **For images containing specific VALUES (numbers, dates, text): return the actual value you see**
- **For images containing VISUAL PATTERNS/TRENDS: describe what you observe in plain English**

For text/table chunks: return cell references or Excel formulas as usual
For image chunks with VALUES: return the actual value from the image
For image chunks with TRENDS/PATTERNS: describe the trend in words

- Use proper Excel syntax with column letters and row numbers (only for non-image data)
- Include sheet name if needed (e.g., Sheet1!A5)
- CRITICAL: Return ONLY the raw JSON object - do NOT wrap it in markdown code blocks, do NOT use ```json``` tags, do NOT add any text before or after the JSON
- Your entire response must be parseable by json.loads() in Python

Format your response as a valid JSON object (NO markdown formatting):

{
  "reasoning": "Explain your analysis here - which chunks you examined, what patterns you found, and how you determined the answer",
  "answer": "Cell reference or Excel formula only (e.g., A1, B5, SUM(B2:B5))"
}

--- (few-shot examples omitted for brevity) ---

Now analyze the question and data, then return ONLY a valid JSON object with "reasoning" and "answer" keys:

Here are the relevant chunks from the spreadsheet(s). CONSIDER ALL RELEVANT CHUNKS PROVIDED:

{relevant_chunks}

REMINDER BEFORE ANSWERING: 
- If the question asks about a TREND, PATTERN, DIRECTION, or VISUAL OBSERVATION from a chart/graph, answer with DESCRIPTIVE WORDS (e.g., "increasing", "declining", "peaked in Q3", "fluctuating pattern")
- If the question asks for a SPECIFIC VALUE from data/cells, answer with cell reference or formula (e.g., "A5", "SUM(B2:B10)")
- NEVER answer "What is the trend?" with "A5" or a cell reference - trends are described in words!

Here is the Question you must answer in json formatting: {task}
\end{lstlisting}
\end{minipage}
\end{center}

\end{document}